# Data-Driven Semi-Supervised Machine Learning with Safety Indicators for Abnormal Driving Behavior Detection

**Yongqi Dong[1*], Lanxin Zhang[1*], Haneen Farah[1], Arkady Zgonnikov[2], and Bart van Arem[1]**

**Abstract**

Detecting abnormal driving behavior is critical for road traffic safety and the evaluation of drivers' behavior. With the advancement of machine learning (ML) algorithms and the accumulation of naturalistic driving data, many ML models have been adopted for abnormal driving behavior detection (also referred to in this paper as "anomalies"). Most existing ML-based detectors rely on (fully) supervised ML methods, which require substantial labeled data. However, ground truth labels are not always available in the real world, and labeling large amounts of data is tedious. Thus, there is a need to explore unsupervised or semi-supervised methods to make the anomaly detection process more feasible and efficient. To fill this research gap, this study analyzes large-scale real-world data revealing several abnormal driving behaviors (e.g., sudden acceleration, rapid lane-changing) and develops a hierarchical extreme learning machine (HELM)-based semi-supervised ML method using partly labeled data to accurately detect the identified abnormal driving behaviors. Moreover, previous ML-based approaches predominantly utilized basic vehicle motion features (such as velocity and acceleration) to label and detect abnormal driving behaviors, while this study seeks to introduce event-level safety indicators as input features for ML models to improve the detection performance. Results from extensive experiments demonstrate the effectiveness of the proposed semi-supervised ML model with the introduced safety indicators serving as important features. The proposed semi-supervised ML method outperforms other baseline semi-supervised or unsupervised methods as far as various metrics are concerned: for example, it delivers the best accuracy at 99.58% and the best F1-score at 0.9913. The ablation study further highlights the significance of safety indicators for advancing the detection performance of abnormal driving behaviors.

**Keywords**

abnormal driving behavior, semi-supervised machine learning, hierarchical extreme learning machines, self-supervised training, safety indicators

Road traffic safety has become a growing concern worldwide. The World Health Organization reported that approximately 1.19 million people die each year in road traffic crashes, with over 30 million suffering non-fatal injuries (*1*). These crashes not only result in disabilities but also cause significant economic loss, reaching as high as 3% of the gross domestic product in some countries. It is alarming that, in most crashes, human factors were identified as contributing factors (*2–4*). This highlights the urgent need to identify abnormal driving behaviors and find ways to prevent or mitigate crashes caused by them.

Driving behavior encompasses various variables and factors, including driving performance, environmental awareness, risk-taking propensity, and reasoning abilities (*5*). Abnormal driving behavior refers to reckless actions that deviate from safe and normal driving, posing risks to the driver, passengers, and other road users, and typically occurs within a short period of time (*6*). Examples of such behavior include excessive speeding, tailgating,

[1]Faculty of Civil Engineering and Geosciences, Delft University of Technology, Delft, The Netherlands
[2]Faculty of Mechanical Engineering, Delft University of Technology, Delft, The Netherlands
[*]These authors contributed equally to this work and should be considered as co-first authors.

**Corresponding Author:**
Yongqi Dong, yongqi.dong@rwth-aachen.de

and erratic lane changes (*7*). These abnormal driving behaviors frequently engender severe traffic altercations, including collisions, crashes, and other minor incidents; this underscores the necessity of addressing and precluding these actions (*6*, *7*). Effective monitoring of abnormal driving behaviors is integral to augmenting driving safety, enhancing driver awareness of driving patterns, and reducing the chances of road crashes.

Machine learning (ML)-based approaches have shown great promise in detecting abnormal driving behaviors. They can learn complex patterns, adapt to changing scenarios, handle large and diverse datasets, and detect unusual behaviors with optimized processes (*8*). However, most of the available studies have adopted fully supervised ML models to do the detection, and few of them explored unsupervised or semi-supervised ML methods. While, in the real world, ground truth labels are sometimes missing or inaccurate, plus labeling large amounts of data is tedious and even dangerous under certain critical situations. Therefore, examining and developing unsupervised or semi-supervised methods is imperative to achieve more feasible and efficient abnormal driving behavior detection.

On the other hand, safety indicators and, particularly, surrogate measures of safety (SMoS) offer a proactive approach to safety evaluation by using proximity measures. Since SMoS do not rely directly on crash data, employing them allows road safety assessment without the need to collect crash data (*9*). As Tarko notes, SMoS facilitates detecting excessive crash risk, better understanding crash-precipitating conditions, and estimating countermeasure efficacy (*10*). By providing insights into potential safety issues, the safety indicators help prioritize improvement efforts. Wang et al. categorize safety indicators into three classes: time-based (e.g., time-to-collision [TTC] and post-encroachment time), deceleration-based (e.g., deceleration rate to avoid a crash), and energy-based (e.g., DeltaV) (*11*). Commonly, these safety indicators are applied in road safety research in combination with thresholds to identify traffic conflicts (*9*, *12*, *13*). There is no doubt that safety indicators can serve as important features in various tasks, for example, in traffic safety assessment and in detecting traffic conflicts. However, for data-driven-based abnormal driving behavior detection, previous studies predominantly employed basic vehicle motion (e.g., speed, acceleration) as features to label and detect abnormal behaviors, and seldom explored the benefits of safety indicators.

To fill the aforementioned research gaps, this study aims to develop a data-driven approach for abnormal driving behavior detection using real-world naturalistic driving data and leveraging semi-supervised ML with self-supervised training to enhance the performance and effectiveness of the detection method. Specifically, this study first analyzes a large-scale dataset, the CitySim dataset, with vivid visualizations, and extracts various abnormal driving behaviors (*14*). Then, the study develops a hierarchical extreme learning machine (HELM)-based semi-supervised ML model using unlabeled data to carry out self-supervised pre-training and leveraging only partly labeled data to fine-tune the model for accurately detecting the identified abnormal driving behaviors. Furthermore, this study conducts a significative ablation study introducing event-level safety indicators as input features for the developed semi-supervised ML model to further improve the detection performance. Extensive experiments verified the proposed method. The proposed semi-supervised HELM model using safety indicators as input features outperforms other baseline models, delivering the best accuracy at 99.58% and the best F1-measure at 0.9913.

**Table 1.** Different Classifications of Abnormal Driving Behaviors

| Chen et al. (*15*) | Kim et al. (*16*) |
|---|---|
| Weaving | Sudden start |
| Swerving | Speeding |
| Sideslipping | Long-standing speeding |
| Fast U-turn | Sudden braking |
| Turning with a wide radius | Sudden overtaking |
| Sudden braking | Sudden changing lanes |
| | Sudden turning |

In short, by filling the research gap and addressing the limitations of existing methods in the literature, this research endeavors to improve road safety and reduce accidents caused by abnormal driving behaviors. It addresses the limitations of traditional supervised approaches and overcomes the scarcity of labeled abnormal driving data. The study analyzes publicly available vehicle trajectory datasets and provides meaningful insights into the identification of abnormal human driving behavior. The conclusions and limitations of this study, as well as future research directions, are discussed at the end of this paper.

## Related Work

Several studies have investigated abnormal driving behaviors, with typical examples of Chen et al. and Kim et al. putting forth definitions reflecting different conceptualizations of driving, as shown in Table 1 (*15*, *16*). Chen et al. emphasized whether the vehicle's location complies with regulations, while Kim et al. prioritized speed modulation (*15*, *16*). In combination, despite these different emphases, both studies suggest that sudden changes in speed or location are key indicators of abnormal driving, regardless of the country where the driving occurs.

Building on this, the current study delineates abnormal driving based on changes in position and velocity, concentrating on behaviors of abrupt starts and emergency braking, as well as rapid and close lane changes. This definition is supported by a comprehensive review of the existing literature, indicating a focus on both the spatial and temporal aspects of driving behavior.

ML-based approaches for detecting abnormal driving behaviors have gained substantial research attention and exhibit robust performance. Both supervised and unsupervised methodologies have been commonly utilized in prior investigations of abnormal driving behavior. Supervised techniques necessitate labeled data during model training, whereby the system ascertains the mapping between inputs and outputs to categorize and predict new data points. For example, Jia et al. devised a model integrating long short-term memory neural network and convolutional neural network (CNN) architectures to pinpoint instances of extreme acceleration and deceleration (*17*). Shahverdy et al. proposed a lightweight one-dimensional CNN (1D-CNN) exhibiting high efficiency and low computational overhead for classifying drivers' behavior into safe, distracted, aggressive, drunk, and drowsy driving (*18*). Ryan et al. simulated an end-to-end model leveraging CNN to compare human and autonomous vehicle driving patterns and adopted a Gaussian-processes-based method to detect driving anomalies (*19*).

Conversely, unsupervised ML techniques entail training models using raw, unlabeled data. This approach is frequently utilized during exploratory phases to derive insights from the dataset. As an illustration, Mohammadnazar et al. developed an architecture leveraging unsupervised ML to quantify driving performance and categorize driving styles across diverse spatial contexts (*5*). Feng et al. proposed a support vector clustering methodology to classify driving styles (e.g., aggressive, normal, defensive) robustly (*20*). Existing literature denotes substantial challenges in accurately identifying anomalies through solely unsupervised ML. As Chandola et al. concluded from their review, unsupervised anomaly detection approaches often demonstrate inferior detection rates and heighten false positive rates on real-world problems (*21*). Correspondingly, Pimentel et al. found, via benchmark assessments, that complete dependency on unsupervised anomaly detection is not recommended, as these techniques fail to detect all anomalies (*22*). Erfani et al. further emphasized that purely unsupervised methodologies lack the learning guidance to precisely differentiate normal from abnormal patterns (*23*). Synthesizing these conclusions, utilizing only unsupervised ML without any labeled data to achieve accurate anomaly detection is hardly possible. Even if viable, it is always possible to further enhance the detection performance of pure unsupervised ML by the use of labeled data. Therefore, there is a research consensus about the necessity of making use of at least partially labeled data to supervise and augment anomaly detection capabilities with semi-supervised ML approaches.

Concerning the features utilized as input for ML models, traditional indicators such as velocity, acceleration, and steering angle have been extensively employed (*17, 24–27*). For example, Lim and Yang considered vehicular data comprising velocity, acceleration, steering angle, and gas pedal position, and leveraged a CNN model to estimate driver drowsiness, workload, and distraction levels (*24*). Li et al. collected lateral vehicle position, steering angle, and speed-related information and implemented a support vector machine model to differentiate between normal and intoxicated driving states (*27*). Incorporating safety indicators (e.g., TTC) into ML-based methods is supposed to be promising for abnormal driving detection but has seldom been investigated. To the best of the authors' knowledge, after extensive review, there is only one relevant study—that of Lu et al., who integrated the representation of TTC together with the driver maneuver profiles into a deep unsupervised learning and clustering method with their proposed transformer-encoder-based model to identify traffic conflicts and non-conflicts (*28*). However, they only investigated situations of one intersection and one roundabout in the U.S., neglecting other various types of driving anomalies, especially those related to highway driving.

Investigating the potential of semi-supervised approaches, which utilize both labeled and unlabeled data, is imperative to enhance abnormal driving behavior detection, yet limited research has explored this direction. By harnessing the additional information from unlabeled data, semi-supervised learning might be able to uncover subtle patterns and behaviors that conventional supervised or unsupervised techniques may overlook. This study endeavors to address this research gap. Moreover, input features are fundamental for ML-based approaches. To enhance detection performance, it is advisable to explore more effective features. In this line of thought, this study seeks to investigate the benefits of event-level safety indicators as input variables and conducts ablation analyses to verify their efficacy in upgrading the detection accuracy.

## Dataset and Data Analysis

### *Description of the Data*

To conduct data-driven research, a high-quality dataset is imperative. After extensive exploration, this study utilizes the CitySim dataset, comprising video-based trajectory data concentrating on traffic safety in the U.S. (*14*).

**Table 2.** Data Sample of the CitySim Dataset

| Feature | Value |
|---|---|
| frameNum | 0 |
| carId | 582 |
| carCenterX (ft) | 462.4 |
| carCenterY (ft) | 184.8 |
| headX (ft) | 469.6 |
| headY (ft) | 184.8 |
| tailX (ft) | 455.3 |
| tailY (ft) | 184.8 |
| Speed (mph) | 39.5 |
| Heading (°) | 180.7 |
| laneId | 10 |

The CitySim dataset encompasses vehicle trajectory information extracted from videos at 30 frames per second captured by 12 drones, spanning six road geometry typologies including freeway segments, signalized intersections, and stop-controlled junctions. The dataset provides precise positional details with measurements accurate to approximately 10 cm in various formats, including pixels, feet, and GPS coordinates, alongside data on velocity, heading angle, and vehicle lane numbers. Table 2 provides the fields of the raw data record and provides one example accessible within the dataset.

Following the research objectives, supplementary features were derived from the CitySim dataset, encompassing, for example, longitudinal acceleration, lateral acceleration, and inter-vehicle distances, which facilitate the calculation of event-level safety indicators. By integrating these computed variables with the original dataset, this study endeavors to strengthen the data foundations necessary for the model. However, the dataset initially still contains noisy and inconsistent data. Rigorous pre-processing techniques were employed to enhance the quality and reliability, ensuring robustness in subsequent analysis and model training. Firstly, entries with missing values and *NULL* were identified and treated using the *dropna* function in the Python *pandas* library, eliminating instances with incomplete information. Then, entries with extreme values, such as distance or speed beyond the normal range, were cleared. For example, negative values in either distance or speed and speed values beyond 100 m/s (360 km/h) are considered extreme values.

Furthermore, a data-smoothing technique with exponential smoothing was applied to attenuate high-frequency noise while preserving the underlying trends and patterns of the data.

Table 3 exhibits examples of the data used after the pre-processing. As illustrated, the data after pre-processing includes features of coordinates, that is, carCenterX and carCenterY, speed, heading angle, and distance. Since carCenterX, carCenterY, speed, and heading angle are provided in the original data, they were the fields used when smoothing the data. The data fields of distance, together with the later introduced longitudinal and lateral acceleration, were calculated after the pre-processing using the relevant fields. For example, distance was calculated using carCenterX and carCenterY of the adjacent two vehicles.

The Methodology section delineates the precise calculations done to derive the additional features from the raw CitySim dataset, including, as well, the selected event-level safety indicators.

## *Abnormal Driving Behaviors Identified in the Dataset*

Based on the classification and definition of abnormal driving behavior in the reviewed literature (check the Related Work section), this section illustrates the specific abnormal driving behaviors observed in the examined CitySim dataset. Each abnormal behavior is associated with one or two indicators, measured or calculated at various locations.

*Rapid Acceleration and Emergency Braking Behavior.* The acceleration data corresponding to each velocity datum in the vehicle trajectory dataset is exhibited in Figure 1.

**Table 3.** Data Examples after Data Pre-Processing

| frameNum | carCenterX (m)* | carCenterY (m)* | Speed (m/s) | Heading (°) | Distance (m)* | Abnormal = 1; normal = 0 |
|---|---|---|---|---|---|---|
| 10 | 53.258 | 32.155 | 14.985 | 359.632 | 0.482 | 1 |
| 1,737 | 251.998 | 27.466 | 11.095 | 359.742 | 131.453 | 0 |
| 1,739 | 248.537 | 31.095 | 12.300 | 359.707 | 128.168 | 0 |
| 1,760 | 251.607 | 27.355 | 11.064 | 359.656 | 131.392 | 0 |
| 11,940 | 128.567 | 31.653 | 16.368 | 359.220 | 0.593 | 1 |
| 11,966 | 127.897 | 31.653 | 16.217 | 359.082 | 0.482 | 1 |
| 11,981 | 127.115 | 31.653 | 16.218 | 358.865 | 0.457 | 1 |
| 12,000 | 126.836 | 31.542 | 16.277 | 358.864 | 0.387 | 1 |

*The original distance measure in feet is converted to meters.

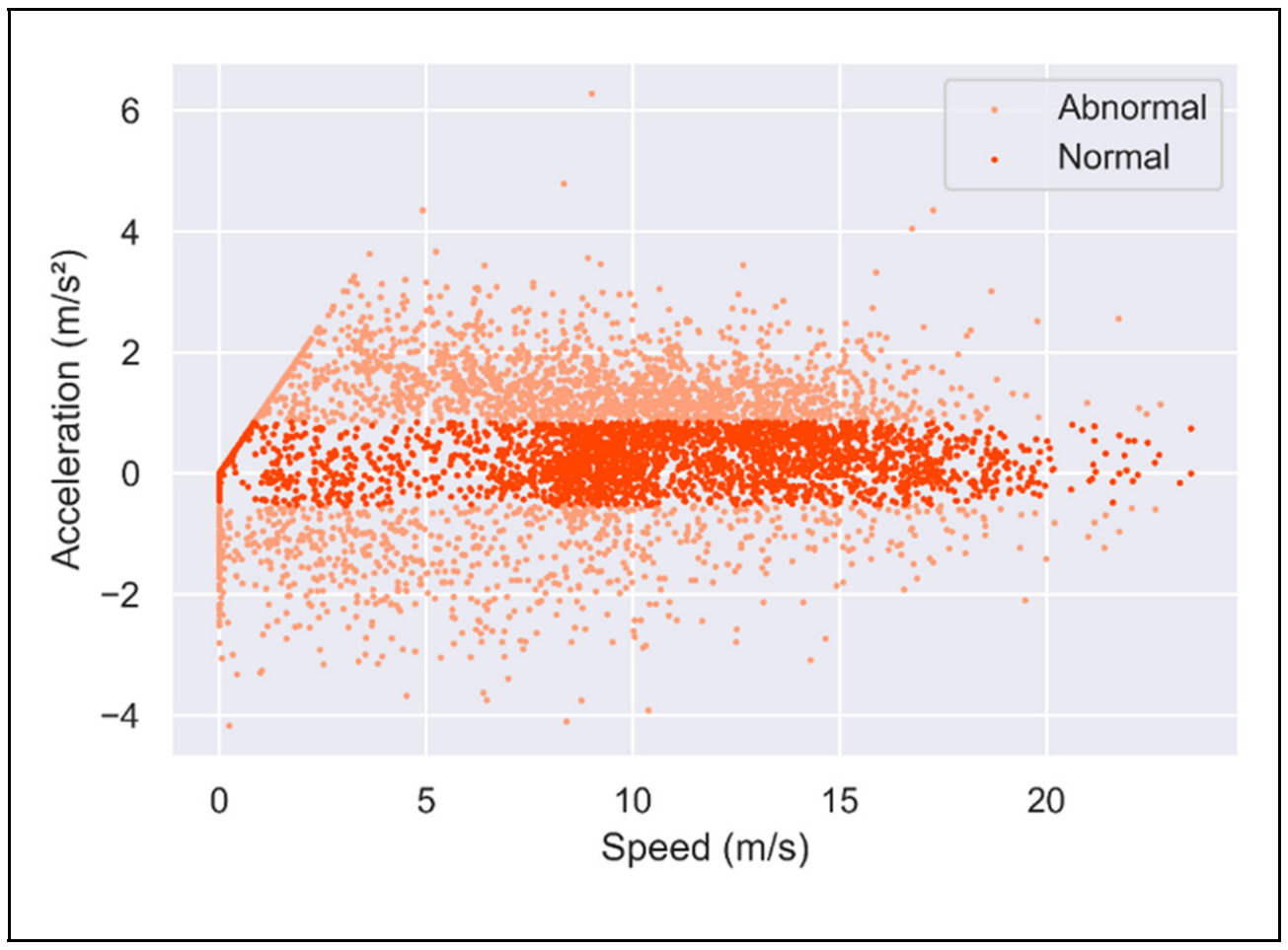

**Figure 1.** Longitudinal acceleration and deceleration scatterplot at different speeds.
*Note*: dark orange dots = normal observations; light orange dots = abnormal observations.

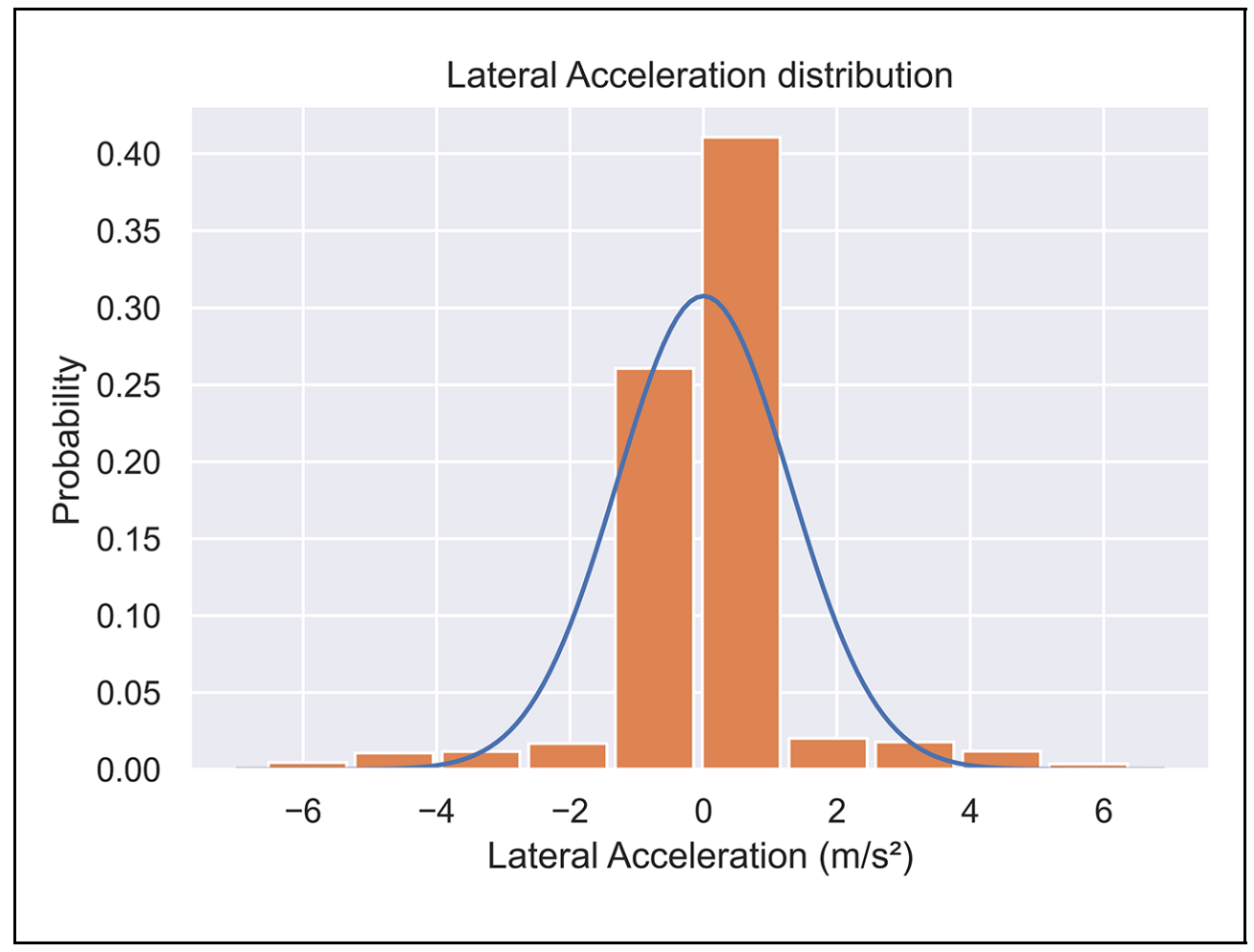

**Figure 2.** Illustration of the distribution of lateral acceleration.

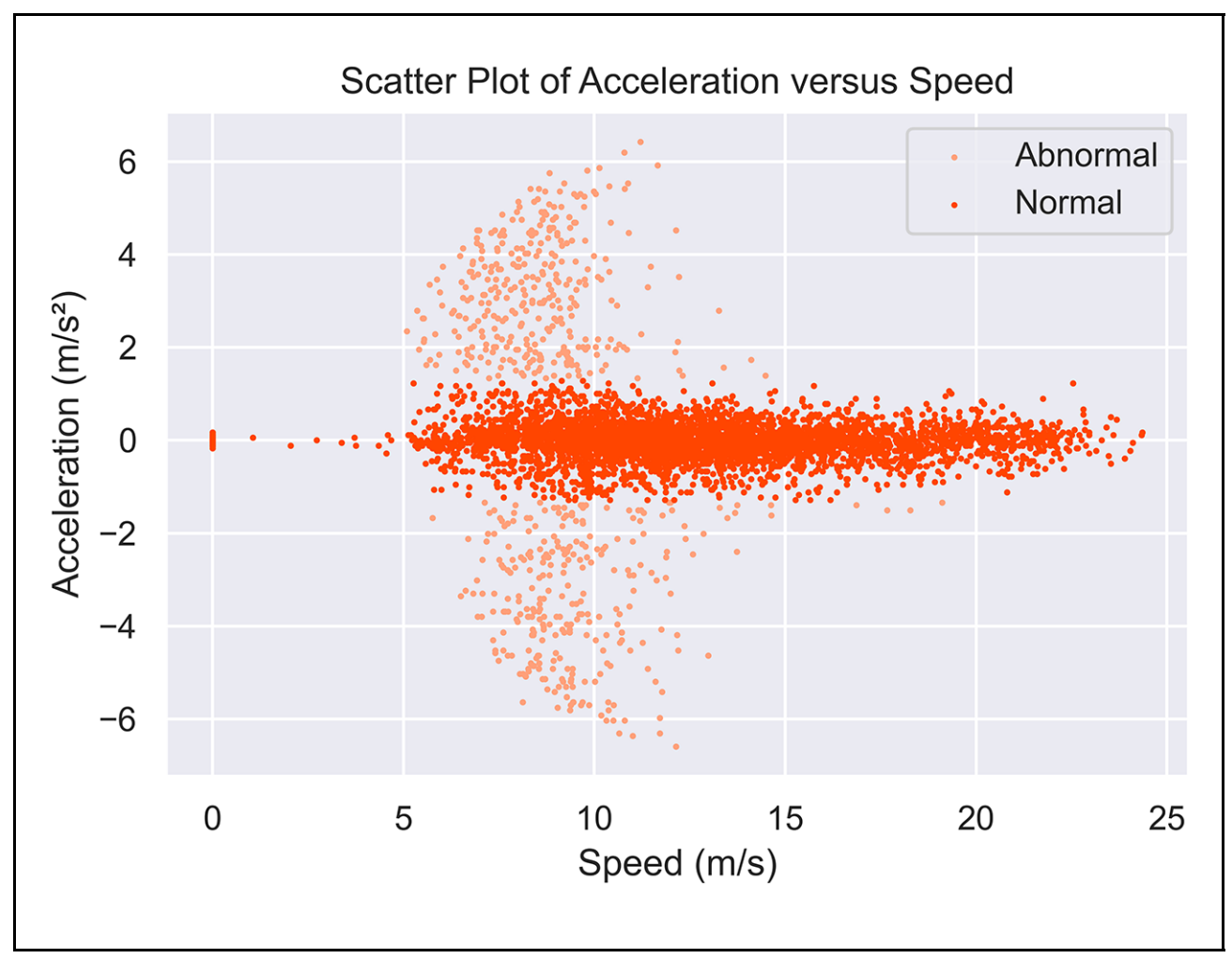

**Figure 3.** Lateral acceleration scatterplot at different speeds.
*Note*: dark orange dots = normal observations; light orange dots = abnormal observations.

Extreme acceleration and deceleration observations can be derived, denoting abnormal maneuvers such as sudden braking or accelerating. Identifying these extreme observations enables the segmentation of abnormal driving behaviors versus normal ones. A specific proportion of extreme acceleration can be pinpointed by statistically scrutinizing all acceleration observations at identical speeds across all journeys. Determining an appropriate ratio to differentiate extreme/abnormal points from normal ones is imperative. A 15% threshold appears sensible based on reiterative experimentation and associated existing research (*17*, *29*).

*Rapid Lane-Changing Behavior.* Rapid lane-changing behavior is characterized by sudden and instantaneous abnormal lateral accelerations that occur for a short duration. In normal driving patterns, vehicles exhibit relatively stable lateral acceleration around zero (as shown in Figure 2). However, abnormal lane-changing behavior manifests an abrupt variation in the vehicle's lateral acceleration.

The majority of vehicles exhibiting lane divergence comportment demonstrate a lateral acceleration bounded by $\pm 1\,m/s^2$, whereby they execute lane diversions seamlessly at a fixed velocity. However, the accelerations of some vehicles appear as outliers in Figure 3. A normal distribution with a mean of 0 and a standard deviation of 1.3 was examined. According to the characteristics of a normal distribution, approximately 68% of data falls within $\pm 1$ standard deviation from the mean. These outliers beyond $\pm 1$ standard deviation from the mean accounted for approximately 32% of the total data points. A ratio of approximately 15% is considered reasonable based on repeated experiments and related research (*17*, *29*). This satisfies the heuristic definition of outliers as observations that differ significantly from most data. Examining outliers based on standard deviation thresholds aligns with statistically grounded techniques for anomaly detection using the sigma principle for normal distributions (*17*, *29*). According to the normal distribution, values greater than $1.3\,m/s^2$ and less than $-1.3 m/s^2$ were used as the filter condition for abnormal instances.

*Close Lane-Changing Behavior.* Close lane-changing behavior is characterized by sudden and instantaneous abnormal lane-changing actions with very short distances from adjacent vehicles that occur for a short duration. Vehicles in normal driving patterns maintain a certain distance between themselves and adjacent lanes. However, during abnormal close lane-changing behavior, there is a significant decrease in the distance between the vehicle and vehicles in the adjacent lanes, indicating a close lane-change. In this study, when the distance between the car performing the lane-changing maneuver and its surrounding vehicles is less than 0.5 m, it is considered severe abnormal driving behavior. In contrast, when the distance is less than 1.0 m but greater than 0.5 m, it is considered weak abnormal driving behavior, as seen in Figure 4.

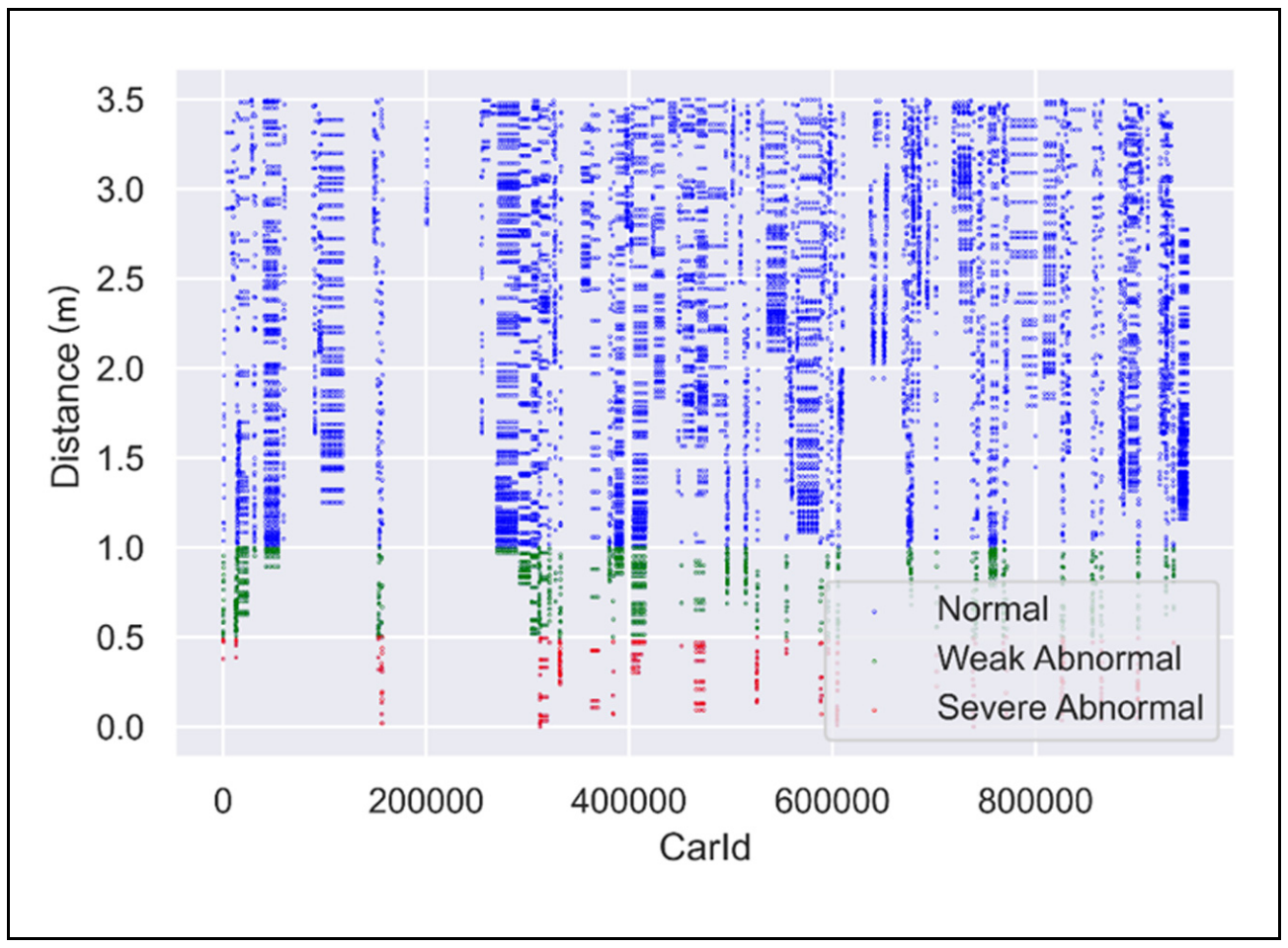

**Figure 4.** Scatterplot of distance during lane-changing for different CarId.

Based on the aforementioned criteria, the labels of the driving data samples were further examined by human experts to remove inaccurate labeling, improving the quality of the finalized labels. Referring to the method adopted by Jia et al., firstly, the data samples during the periods with large longitudinal accelerations and decelerations, large lateral accelerations, and extreme close distances, that is, grey area data, were selected to be checked and verified (*17*). By observing the changes in the distribution of the extreme longitudinal acceleration and deceleration data points, lateral accelerations, and distance-changing dynamics, the human expert combined these observations with their knowledge and experience to verify the labels. If the human expert was not certain with high confidence about the labeling for the data sample, that specific data sample was removed. It should be noted that this human-expert-examination-based verification method may only correct the labels of false alarms, to the degree that is possible through examining kinematic variables, and will not correct missed abnormal instances.

## Methodology

This section first introduces event-level safety indicators, especially the adopted two-dimensional TTC (2D-TTC) (*30*, *31*). Two ML models, that is, isolation forest, and robust covariance, are then presented as baseline methods for comparison. Finally, a customized semi-supervised mode, HELM, is proposed and explained in detail.

### *Safety Indicators*

In the literature, several safety indicators were developed and introduced—a comprehensive overview can be found in Nickloaou et al. and Arun et al. (*9*, *32*). One of the most popular and commonly used safety indicators is the TTC is a time-based proximity measure. TTC is defined as the time required for two road users, on a collision course, to collide if no evasive action is taken, and this can be and is generally computed continuously (*33*). Its simplistic form is when road users' speed and path are assumed to remain unchanged (*34*). For example, the TTC value, for a car-following situation, assuming motion prediction with constant speed, is calculated as:

$$\mathrm{TTC} = \frac{D}{v_1 - v_2} \quad (1)$$

where

$D$ = the distance between the following and the leading vehicle,

$v_1$ = the speed of the following vehicle, and

$v_2$ = the speed of the leading vehicle.

Over the years, several studies have further extended the TTC safety indicator. For example, time exposed TTC and time integrated TTC were introduced by Minderhoud and Bovy to measure the risk associated with the duration of dangerous driving conditions (*35*). The modified TTC (MTTC) proposed by Ozbay et al. provides an alternative way to calculate TTC at each instant; for example, in a car-following traffic scenario, by considering the accelerations of both the lead and following vehicles (*36*). Other approaches involve incorporating site-specific motion patterns of road users and calculating TTC with respect to the distribution of possible trajectories (*37*, *38*). In this study, a TTC-based safety indicator, 2D-TTC, was implemented as an input feature, which can capture proximities of vehicles' movements and interactions in a plane in various traffic scenarios besides a car-following scenario (*30*, *31*). The illustration of 2D-TTC is demonstrated in Figure 5. 2D-TTC is calculated as follows:

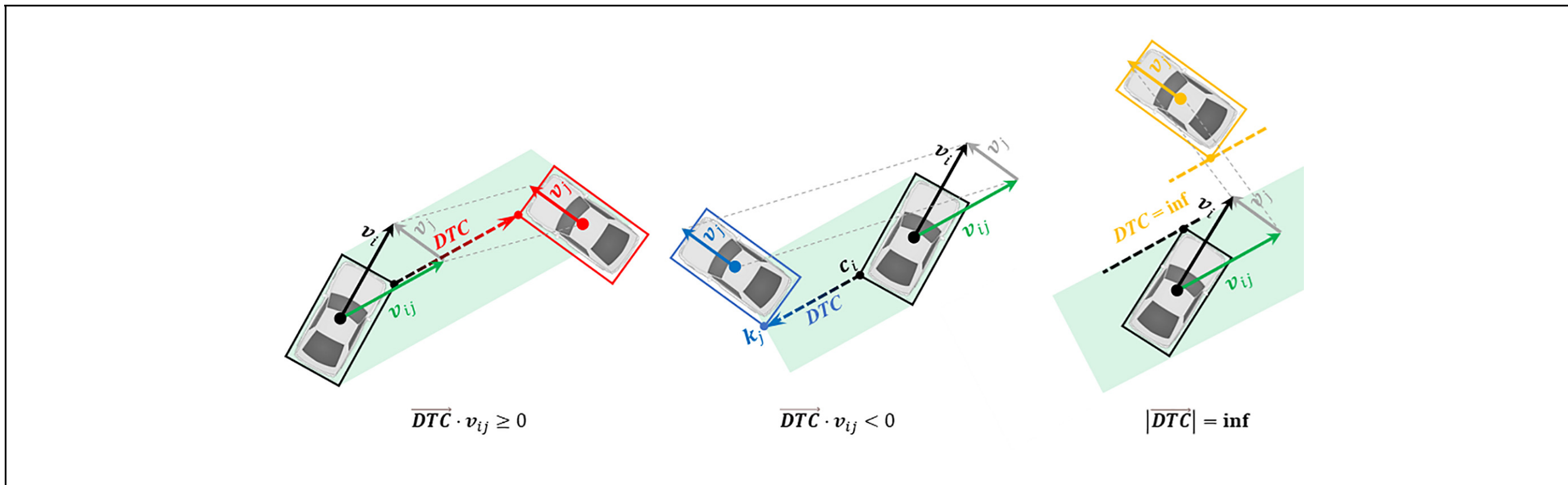

**Figure 5.** Illustration of two-dimensional time-to-collision.
*Source*: adapted from Jiao et al. (*31*).
*Note:* DTC = distance-to-collision.

$$\text{2D-TTC} = \begin{cases} \dfrac{|\overrightarrow{\boldsymbol{DTC}}|}{|\boldsymbol{v}_i - \boldsymbol{v}_j|}, \text{if the direction of } \overrightarrow{\boldsymbol{DTC}} \text{ is the same as } \boldsymbol{v}_{ij} = (\boldsymbol{v}_i - \boldsymbol{v}_j) \\ inf, if |\overrightarrow{\boldsymbol{DTC}}| = \text{inf } OR \text{ if the direction of } \overrightarrow{\boldsymbol{DTC}} \text{ is opposite with } \boldsymbol{v}_{ij} = (\boldsymbol{v}_i - \boldsymbol{v}_j) \end{cases} \quad (2)$$

where

$|\overrightarrow{\boldsymbol{DTC}}|$ = the distance-to-collision (the minimum distance between the bounding boxes of target vehicle $i$ and another, interacting, vehicle $j$ along their relative speed $\boldsymbol{v}_{ij} = (\boldsymbol{v}_i - \boldsymbol{v}_j)$ direction),
$\boldsymbol{v}_i$ = the speed of the target vehicle $i$, and
$\boldsymbol{v}_j$ = the speed of the interacting vehicle $j$.

If the relative movement of target vehicle $i$ and interacting vehicle $j$ decreases the distance-to-collision, they are approaching each other, and a potential collision exists. Otherwise, the vehicles are moving away from each other and no potential collision exists. For more detailed information about the demonstration and calculation of the adopted 2D-TTC, the reader is advised to refer to Jiao et al. (*31*).

In general, according to the literature, only encounters with a minimum TTC below 1.5 s are deemed critical, with trained observers consistently applying this threshold in practice (*39*). This study explores the effects of the input feature 2D-TTC on the detection performance of abnormal driving behavior. Specifically, the vehicle angle in the dataset decomposes each vehicle's velocity into *x*-*y* coordinate components, yielding velocity vectors based on the dataset parameters. 2D-TTC is then calculated per these velocity vectors and the corresponding distance along the same direction. This approach highlights how 2D-TTC can be computed from the raw dataset by leveraging the vehicle angle data to obtain velocity vectors in coordinate space. The derived 2D-TTC is analyzed and integrated with input features such as position, speed, and acceleration to evaluate abnormal driving behavior detection performance using the given dataset.

## *Baseline Models*

Isolation forest and robust covariance are selected as two baseline methods, considering their interpretability, effectiveness, and broad utilization in various domains.

The isolation forest, initially developed by Liu et al., constitutes an effective algorithm typically utilized for data anomaly detection (*40*). The isolation forest algorithm is based on the principle that anomalous data points are more readily separable from the majority of normal samples. To isolate an abnormal data point, the algorithm iteratively generates partitions of the sample by randomly selecting a feature attribute and subsequently randomly choosing a split value within the permissible minimum and maximum values for the selected feature attribute. Through recursive binary partitioning, data points that require fewer splits to become isolated are deemed more anomalous.

The isolation forest algorithm capitalizes on the premise that anomalies are few and different from the rest of the data, and thereby manifest topological shorter path lengths from the root to the external node (leaf), (which is elucidated by averaging this value across the trees) when random partitioning is employed. Therefore, it leverages an ensemble of isolation trees generated through such recursive random partitioning to identify anomalies, with shorter average path lengths corresponding to greater anomaly scores.

In practice, the isolation forest anomaly detection algorithm involves two primary phases. Firstly, a collection of isolation trees (iTrees) is constructed utilizing recursive partitioning on a training dataset. During recursive partitioning, splits are performed by randomly selecting an attribute and random split value to isolate a data point. Secondly, each instance in the test set is propagated through the ensemble of iTrees and assigned an "anomaly score" based on the average path length for that instance across the iTrees. Shorter average path lengths correspond to fewer partitions required to isolate the instance, indicating more anomalous behavior and higher anomaly scores. After computing anomaly scores for all test instances, those data points with a score exceeding a predefined threshold specific to the domain can be classified as anomalies.

The robust covariance estimation algorithm presupposes that normal data points exhibit a Gaussian distribution, and, accordingly, approximates the morphology of the joint distribution (namely, estimates the mean and covariance of the multivariate Gaussian distribution) (*41*).

In statistical analysis, the deviation can be measured by the Z-score. The generalization of the Z-score for a point $x_i$ in the case of a $p$-dimensional multi-variate probability distribution with some mean $\mu$ and covariance matrix $\Sigma$ is known as Mahalanobis distance $d_i$, which is given by:

$$d_i = \sqrt{(x_i - \mu)^T \Sigma^{-1} (x_i - \mu)} \quad (3)$$

It is based on the premise that outliers increase the values (entries) in $\Sigma$, thereby making the data dispersion appear more extensive. Consequently, $|\Sigma|$ (the determinant) will also be larger, which could theoretically decrease if extreme samples are removed. Rousseeuw and Van Driessen devised a computationally efficient algorithm capable of furnishing robust covariance approximations (*42*). The approach assumes that, at minimum, $h$ of the $n$ samples are "normal" ($h$ denoting a hyperparameter). The algorithm begins with $k$ arbitrary samples containing ($p$ + 1) points. For each $k$ sample, $\mu$, $\Sigma$, and $|\Sigma|$ are estimated, the distances are computed and sorted in ascending order, and the smallest $h$ distances are employed to update the estimates. In their original publication, the process of computing distances and revising the estimations of $\mu$, $\Sigma$, and $|\Sigma|$ is entitled a "C-step" whereby two such increments are typically sufficient to identify effective candidates (for $\mu$ and $\Sigma$) among the $k$ arbitrary samples. In the succeeding step, a subset of magnitude $m$ with the lowest $|\Sigma|$ (the optimal candidates) is contemplated for computation until convergence, and the sole estimate whose $|\Sigma|$ is minimal is furnished as output.

Note that, although isolation forest and robust covariance are usually considered unsupervised ML approaches, in this study only normal data samples are input to train them; thus, in this study, they can be regarded as semi-supervised approaches and are comparable with the proposed semi-supervised ML method.

### HELM-Based Semi-Supervised ML

The HELM algorithm, originally proposed by Tang et al., constitutes an advanced extension of the extreme learning machine (ELM) algorithm that can enhance performance in both training speed and generalization capability (*43*). This approach integrates a feed-forward neural network structure with multiple latent layers, and it operates through two primary steps: unsupervised feature representation and supervised feature classification. In the initial step, HELM is intended to ascertain a sparse encoder in an unsupervised manner, which transforms the raw input into superior-level representation. The encoder is structured with multiple latent layers which are processed sequentially, with each layer building on the previous one to capture increasingly abstract features of the data. The second step involves using these learned features for supervised classification or approximation tasks. By leveraging the rich, hierarchical features extracted in the first step, HELM aims to achieve effective and accurate predictions. This two-step process enables HELM to combine the advantages of both unsupervised and supervised learning, resulting in improved overall performance.

Given a training set with $N$ samples, indicated by $(X_i, Y_i)(X_i \in R^n, Y_i \in R^t, i = 1, 2, 3, \ldots, N)$, where $X_i$ and $Y_i$ denote the feature representation and the targeted output of the $i$ th sample, respectively. Suppose the encoder consists of $K$ hidden layers, each with $L_i (1 \leq i \leq K)$ neurons. The output $O = [o_1, o_2, \ldots, o_N]^T$ can be expressed as:

$$\sum_{i=1}^{K} \beta_i g(W_i \cdot x_j + b_j) = o_j, j = 1, 2, \ldots, N \quad (4)$$

where

$g(\cdot)$ = the activation function,

$\beta_i$ = the output weight,

$W_i$ = the input weight, and

$b_j$ = the bias.

Ideally, there should be:

$$\sum_{j=1}^{N} ||o_j - Y_j|| = 0 \quad (5)$$

This implies there exist weights $\beta_i$, $W_i$, and biases $b_i$ such that:

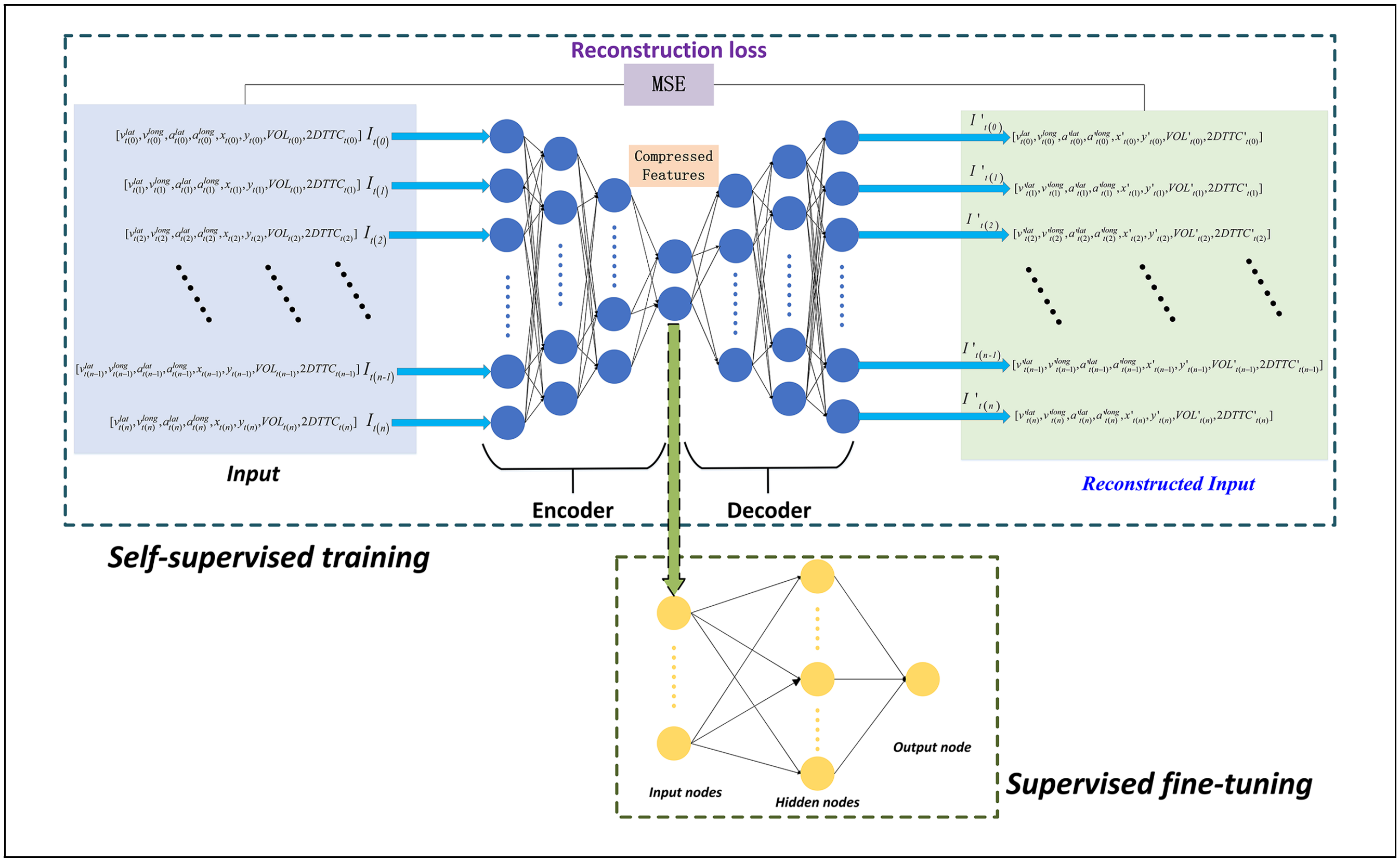

**Figure 6.** The framework of hierarchical extreme learning machine (HELM)-based semi-supervised machine learning method.
*Note:* MSE = mean square error.

$$\sum_{i=1}^{k} \beta_i g\left(W_i \cdot x_j + b_j\right) = Y_j, \quad \text{for } j = 1, 2, \ldots, N \tag{6}$$

In matrix form, this can be represented by:

$$H\beta = Y \tag{7}$$

where
$H$ = the output of the hidden layer node,
$\beta$ = the output weight, and
$Y$ = the desired output.

$$H(W_1, W_2, \ldots, W_K, b_1, b_2, \ldots, b_K, x_1, x_2, \ldots, x_N) = \begin{bmatrix} g_1(X_1) & \cdots & g_{K_1}(X_1) \\ \vdots & \ddots & \vdots \\ g_1(X_N) & \cdots & g_{K_1}(X_N) \end{bmatrix} \tag{8}$$

To train a single hidden layer ELM neural network is equivalent to obtaining $\hat{\beta}$ such that:

$$||H\hat{\beta} - Y|| = \min_{\beta} ||H\beta - Y|| \tag{9}$$

When choosing the mean square error as the measure, this formula is equivalent to minimizing the following loss function:

$$\text{Loss} = \sum_{j=1}^{N} \Big( \sum_{i=1}^{K} \beta_i g\left(W_i \cdot x_j + b_i\right) - Y_j \Big)^2 \tag{10}$$

Traditional ELMs allow the weights β and the deviations between the latent layers and the inputs to be set arbitrarily, drawn from any distribution. This flexibility means that the learning process primarily adjusts these weights to find the optimal connections between the latent layers and the output. However, standard ELMs can be limited in their ability to effectively process complex data, even with many hidden nodes.

In this study, the customized HELM was introduced to address this limitation, which stacks multiple layers of ELM to create a deeper and more profound structure. This hierarchical approach enhances the model's ability to capture intricate data patterns. The proposed HELM-based semi-supervised learning consists of two phases: 1) self-supervised training for feature learning, where the model extracts and learns useful features from the data in an unsupervised manner, and 2) supervised fine-tuning, where the model is further optimized using samples of labeled data to improve its performance on the abnormal driving behavior detection task, as visualized in Figure 6.

The HELM model is initially trained purely self-supervised on normal data samples exclusively, with all anomalous examples excluded from this training set. During this phase, by minimizing a reconstruction error loss function, the stacked ELM autoencoder layers learn to capture the most salient features of the input data that represent its intrinsic normal characteristics. These extracted feature representations can encapsulate the essential properties of standard normal behavior. Subsequently, the learned feature embeddings are transferred to a one-class classifier, which undergoes further supervised training to obtain a decision threshold $\tau$. This threshold calibration phase notably utilizes an unseen validation dataset containing only normal data samples. Withholding this validation set during ELM feature learning prevents overfitting the threshold to any potential anomalies in the original training data. Overall, this staged approach enables robust unsupervised feature extraction from normal data, followed by supervised threshold tuning to facilitate effective anomaly detection. Usually, a good threshold $\tau$ can be expressed by:

$$\tau = \gamma \cdot \text{percentile}_p(|1 - \boldsymbol{Y}^{valid}|) \tag{11}$$

where

$\boldsymbol{Y}^{valid}$ = the output of the one-class classifier, and

$\text{percentile}_p$ = a function of the $pth$ percentile with hyperparameters $p$ and $\gamma \geq 0$.

Finally, in the deployment phase, newly observed data samples are propagated through the trained HELM model to obtain the corresponding outputs from the one-class classifier. These outputs, denoted by $\boldsymbol{Y}^{test}$, are compared against the decision threshold $\tau$ established during the training process. Recall that this threshold was calibrated on the separate validation dataset to avoid overfitting. The label assignment for each new test sample is then determined by thresholding its one-class output as follows:

$$Label_{\boldsymbol{Y}^{test}} = sgn(\tau - |1 - \boldsymbol{Y}^{test}|) \tag{12}$$

In summary, the trained HELM model generates layered feature representations of newly observed test data in a purely data-driven manner. Anomalies can be effectively detected by propagating these examples through the model and comparing the resulting one-class classifier decisions with the calibrated threshold $\tau$. This approach benefits from the model's unsupervised learning of salient features from normal training data, and the deep HELM architecture captures robust intrinsic representations of standard normal behavior. By thresholding the one-class outputs relative to $\tau$, deviations from the learned normality are identified during deployment. Overall, this framework provides a self-supervised feature-learning mechanism to represent normal data and a thresholding technique for effective anomaly detection in practice. The model framework of the HELM-based semi-supervised ML method is delineated in Figure 6.

## Experiment and Results

### Dataset Arrangement

This study carries out comprehensive experiments to assess the performance of various models and the impact of different input feature conditions on the detection of abnormal driving behavior. Initially, the built training dataset contained 290,690 instances, which included noisy and inconsistent data. In this study, several techniques were employed to address these issues, such as utilizing the *dropna* function in the *pandas* library to eliminate instances with *NULL*, missing, and blank values, as well as refining the original data by employing smoothing techniques to attenuate noise.

The dataset itself includes the following features: frameNum, carId, carCenterX (ft), carCenterX (m), carCenterY (ft), carCenterY (m), headX (ft), headY (ft), tailX (ft), tailY (ft), speed (mph), speed (m/s), heading, and laneId, as shown in Table 2. Next, the time interval was determined by calculating the difference in timestamp values using frameNum between adjacent later samples and their corresponding former ones. Based on this, the speed together with acceleration (both longitudinal and lateral) for each vehicle was computed. Subsequently, using frameNum as the index, the distances and 2D-TTC between all relevant vehicles at the same timestamp were calculated. As the quantity of normal driving data samples is far beyond the abnormal ones, to balance the quantity of abnormal and normal data samples, this study sampled the normal driving samples. In the end, the examined dataset comprised a total of 23,605 samples, consisting of 12,125 normal instances and 11,480 anomaly instances. All anomaly instances were utilized for testing and 3,638 normal instances were adopted for testing. As anomaly instances are more critical, this study examined more anomaly instances in the estimation of model performance.

### Evaluation Metrics

Various metrics are adopted to evaluate the overall performance of the selected model, and the discrimination evaluation of the optimal model can be defined based on the confusion matrix, as shown in Table 4 (*44*).

In binary classification, one class constitutes the positive class, whereas the other delineates the negative class. The positive class epitomizes the events the model endeavors to detect, that is, abnormal driving in this study, while the negative class constitutes other contingencies, that is, normal driving in this study. True positive (TP) and true negative (TN) denote the quantity of accurately classified positive and negative instances. In this study, TP represents the correctly detected abnormal driving behavior data sample, and TN constitutes the accurately detected normal driving samples. On the other hand, false positive (FP) and false negative (FN) represent the number of misclassified positive and negative instances, meaning incorrect detection of abnormal driving behavior/normal driving behavior instances. Accuracy, precision, and recall were computed based on these four terms.

**Table 4.** Confusion Matrix and the Corresponding Array Representation

| | Actual positive class | Actual negative class |
|---|---|---|
| Predicted positive class | True positive | False negative |
| Predicted negative class | False positive | True negative |

Accuracy refers to the proportion of true results among the total number of cases examined:

$$\text{Accuracy} = \frac{\text{TP} + \text{TN}}{\text{TP} + \text{TN} + \text{FP} + \text{FN}} \tag{13}$$

Precision is utilized to gauge the accurate prediction of positive patterns among the total predicted patterns in a positive class:

$$\text{Precision} = \frac{\text{TP}}{\text{TP} + \text{FP}} \tag{14}$$

Another widely utilized measure is recall, which accounts for the proportion of actual positives that are correctly classified:

$$\text{Recall} = \frac{\text{TP}}{\text{TP} + \text{FN}} \tag{15}$$

The F1-score is a measure combining and balancing precision and recall, and it is defined as the harmonic mean of precision and recall:

$$F1 - \text{score} = 2 \times \frac{\text{precision} \times \text{recall}}{\text{precision} + \text{recall}} \tag{16}$$

Finally, the true positive rate (TPR) and false positive rate (FPR) are also examined as evaluation metrics. As indicated in their names, TPR and FPR are calculated as:

$$\text{TPR} = \frac{\text{TP}}{\text{TP} + \text{FN}} \tag{17}$$

$$\text{FPR} = \frac{\text{FP}}{\text{FP} + \text{TN}} \tag{18}$$

**Table 5.** Input Features in Different Settings

| Experimental Setting | Input features |
|---|---|
| 1 | coordinates/velocity/angle |
| 2 | coordinates/velocity/angle/acceleration/distance |
| 3 | coordinates/velocity/angle/2D time-to-collision |

## *Ablation Study of Features*

Three experimental settings with distinct feature representations are designed to evaluate the impact of input information on model performance. As illustrated in Table 5, Setting 1 utilizes only the raw coordinates, velocity, and vehicle angle features inherently present in the dataset. Setting 2 augments Setting 1 by incorporating two additional engineered features of lateral acceleration and inter-vehicle distance. Finally, Setting 3 further supplements Setting 2 by including the 2D-TTC feature capturing temporal proximity. By comparing results between these controlled settings, the incremental value of providing basic motion features (Setting 2) and safety indicators, that is, 2D-TTC (Setting 3), over the raw dataset (Setting 1) can be quantified. The proposed three experimental settings serve to illustrate the effect of step-wise enriching the feature space on the learning capabilities of the model under controlled conditions.

## *Results and Comparison*

The testing results of the proposed HELM model, together with the two baselines, are illustrated in Table 6, as well as Figures 7 to 9. In general, the HELM model outperforms robust covariance and isolation forest, with the best variant delivering the best accuracy at 99.58% and the best F1-score at 0.9913.

Experiments across three experimental settings demonstrate enhanced abnormal driving behavior identification capabilities by incorporating the safety indicator of 2D-TTC. Furthermore, the proposed semi-supervised HELM model achieves consistently superior performance compared with the alternative baseline models in all three experimental settings.

In the baseline Setting 1, with only raw coordinates, velocity, and angle serving as the input features, the HELM model attains an accuracy of 0.9471. Then, augmenting with acceleration and inter-vehicle distance features, in Setting 2, the accuracy of HELM is improved to 0.9614. Notably, by the further inclusion of the adopted 2D-TTC safety indicator in Setting 3, the accuracy of HELM is dramatically enhanced to 0.9958, alongside near-perfect scores for precision (0.9963), recall (0.9983), F1-score (0.9913), and FPR (0.0118). This underscores

**Table 6.** Comparison Results under Different Settings

| Setting | Accuracy | Precision | Recall | F1-score | FPR | TPR |
|---|---|---|---|---|---|---|
| Robust covariance | | | | | | |
| 1 | 0.3337 | 0.7628 | 0.1779 | 0.3735 | 0.1745 | 0.1779 |
| 2 | 0.3348 | 0.7702 | 0.1767 | 0.3762 | 0.1663 | 0.1767 |
| 3 | 0.9570 | 0.9487 | 0.9973 | 0.9028 | 0.1701 | 0.9973 |
| Isolation forest | | | | | | |
| 1 | 0.5789 | 0.8766 | 0.5185 | 0.4680 | 0.2303 | 0.5185 |
| 2 | 0.4387 | 0.8673 | 0.3080 | 0.4219 | 0.1487 | 0.3080 |
| 3 | 0.9615 | 0.9517 | **1.0000** | 0.9131 | 0.1600 | **1.0000** |
| HELM | | | | | | |
| 1 | 0.9471 | 0.9349 | **1.0000** | 0.8766 | 0.2196 | **1.0000** |
| 2 | 0.9614 | 0.9561 | 0.9949 | 0.9144 | 0.1440 | 0.9949 |
| 3 | **0.9958** | **0.9963** | 0.9983 | **0.9913** | **0.0118** | 0.9983 |

*Note*: FPR = false positive rate; HELM = hierarchical extreme learning machine; TPR = true positive rate.

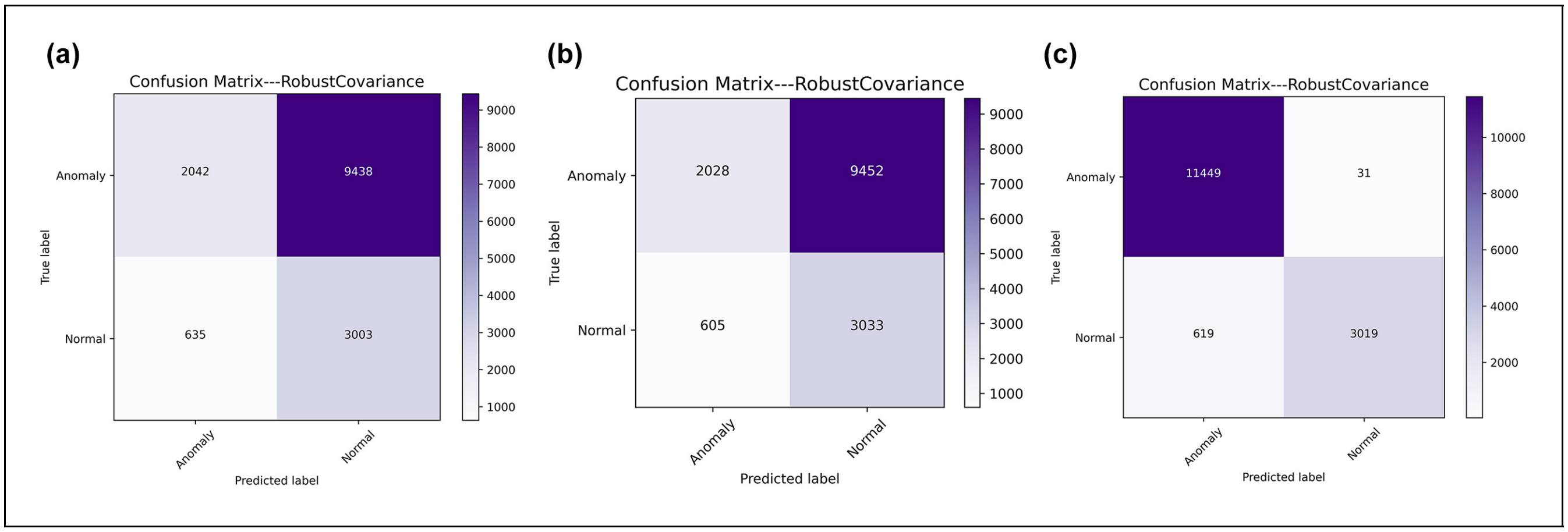

**Figure 7.** Robust covariance performance under (*a*) Setting 1, (*b*) Setting 2, and (*c*) Setting 3.

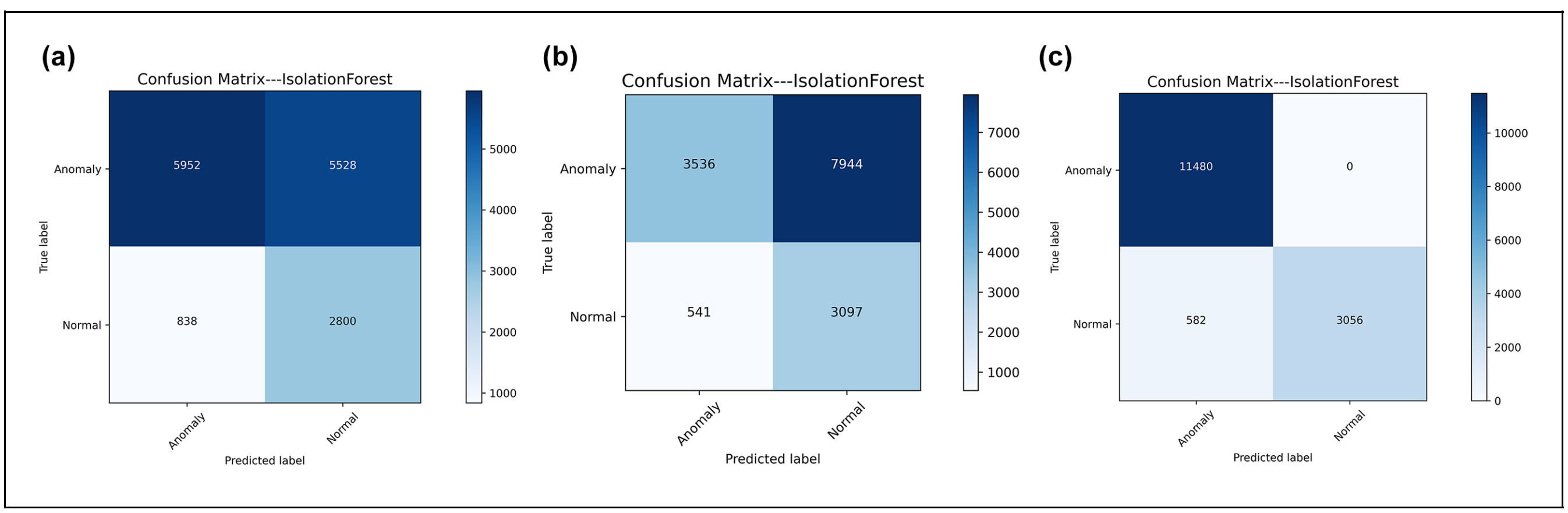

**Figure 8.** Isolation forest performance under (*a*) Setting 1, (*b*) Setting 2, and (*c*) Setting 3.

the outstanding value of 2D-TTC as an important spatial-temporal feature for this task.

Similarly, unsupervised models (which work in a semi-supervised way in this study) exhibit substantial gains when endowed with 2D-TTC. For instance, the precision and recall of robust covariance are improved by over 20%, while the accuracy and F1-score of isolation forest are increased by 5% and 10%, respectively. Nevertheless, the semi-supervised HELM approach outperforms these two baseline models across all metrics except for TPR.

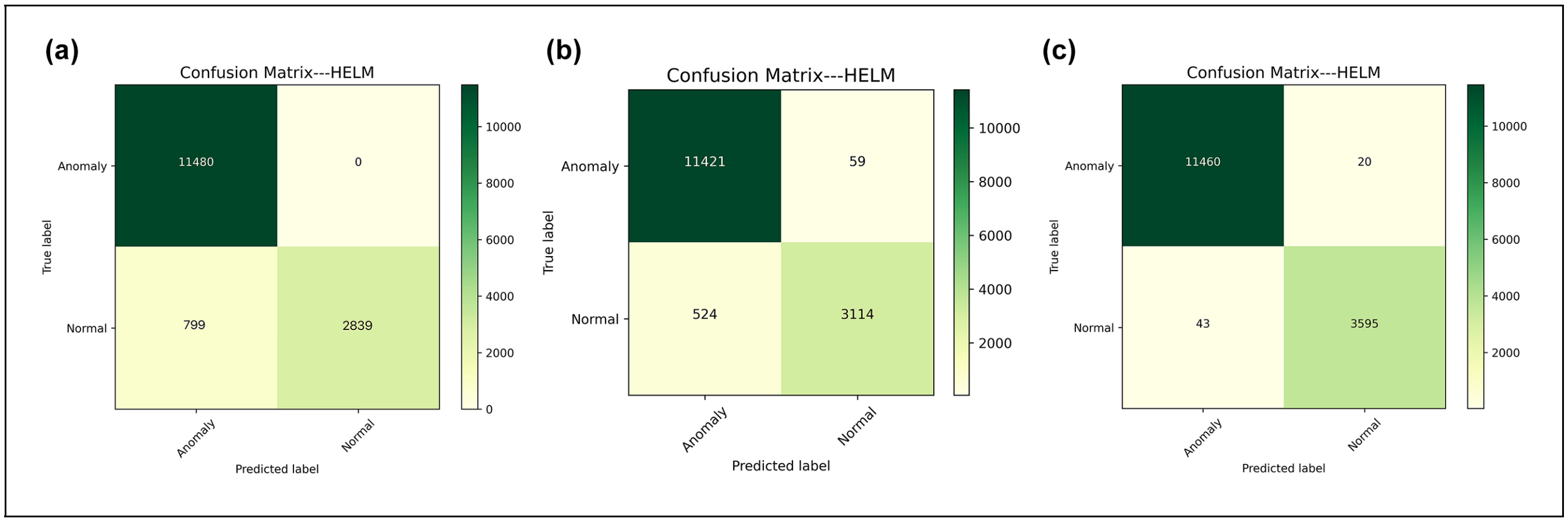

**Figure 9.** Hierarchical extreme learning machine (HELM) performance under (*a*) Setting 1, (*b*) Setting 2, and (*c*) Setting 3.

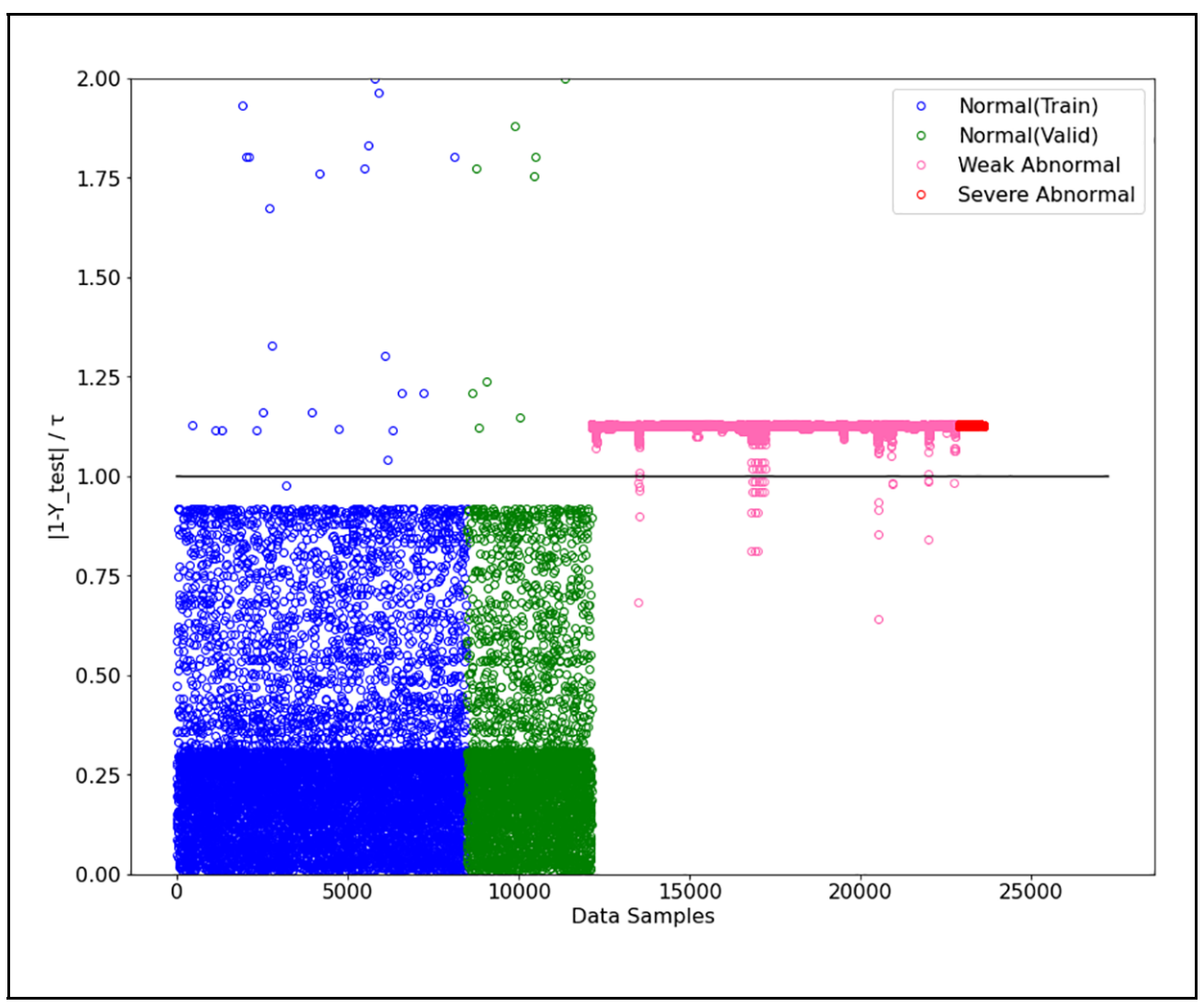

**Figure 10.** Scatter visualization of the result obtained by semi-supervised hierarchical extreme learning machine (HELM).

Finally, scatter visualization of the results obtained by the proposed semi-supervised method using HELM is provided in Figure 10. From the visualization, it is further demonstrated that HELM can distinguish between normal and abnormal driving behaviors. However, it cannot tell the severe abnormal apart from the weak abnormal instances, as the values of their $|1 - Y^{test}|/\tau$ are similar. How to distinguish the severity of abnormal driving behavior using semi-supervised ML can be an interesting future research direction.

In summary, augmenting the feature space with the adopted safety indicator, that is, 2D-TTC, consistently improves the anomaly detection capabilities across models. The HELM framework integrating 2D-TTC markedly surpasses other baseline models, demonstrating the advantages and superiority of the proposed semi-supervised learning method together with the spatial-temporal feature engineering for anomalous driving behavior detection.

## Conclusion and Future Work

This study presents a semi-supervised ML framework leveraging event-level safety indicators to enhance abnormal driving behavior detection. A large-scale real-world naturalistic driving dataset was analyzed and various abnormal driving behaviors were revealed and categorized in this study. The HELM model was proposed, which harnesses unlabeled data for self-supervised pre-training and partially labeled data for fine-tuning. The 2D-TTC safety indicator was introduced as an important feature, with experiments demonstrating that integrating 2D-TTC significantly improves the detection accuracy by over 5% for all the tested models compared with baseline experimental feature settings.

By training on unlabeled data, and employing only a small sample of labeled data for fine-tuning, the proposed semi-supervised approach achieved competitive performance while reducing dependency on fully labeled datasets, making it well-suited for real-world applications with limited labeled data. Notably, the incorporation of event-level safety indicators, in this case 2D-TTC, greatly enhanced the model performance. These compelling results underscore the critical value of safety indicators in effectively detecting abnormal driving behaviors across diverse ML algorithms. This fusion of semi-supervised ML and utilization of safety indicators as input features showcases the potential for advancing abnormal driving behavior detection capabilities, with significant implications for safety-oriented research and evaluations. To further upgrade the detection performance, future studies could explore other and more advanced safety indicators.

Furthermore, the current study focuses on detection, future research should explore predictive capabilities to enable earlier identification of impending abnormal behaviors before manifestation. This involves inputting multi-step time-series driving data and computing the features (e.g., TTC, 2D-TTC, MTTC) over a continuous duration period based on observed historical driving behavior data to predict the status of the next time step or the next few time steps. Additionally, incorporating motion prediction (e.g., for more accurate TTC calculation) and adoption of driving risk field related metrics, such as the human perceived driver's risk field and the probabilistic driving risk field, together with developing techniques to extract robust spatial-temporal patterns as model inputs, could further enhance the detection and prediction performance (*45*, *46*).

Lastly, concerning other limitations, the adopted dataset encompassed only three abnormal driving behavior types in this study. Future research should incorporate an expanded diversity of abnormal driving behaviors and more advanced safety indicators to enrich the understanding and identification of anomalies. Additionally, ground truth labels are the prerequisite for evaluating the model performance. The current human-expert-examination-based verification method adopted in this paper cannot detect missed abnormal driving behavior instances but it may correct possible false alarms to upgrade the label quality. It is suggested to adopt more advanced approaches to obtain and verify high-quality ground truth labels, for example, employing online crowdsourcing with multiple experts, and using more comprehensive datasets with corresponding video recordings, as well as incorporating fine-labeled accident data from road authorities.

## Author Contributions

The authors confirm contribution to the paper as follows: study conception and design: Y. Dong, L. Zhang; data collection: L. Zhang, Y. Dong; analysis and interpretation of results: L. Zhang, Y. Dong, H. Farah, A. Zgonnikov, B. van Arem; draft manuscript preparation: Y. Dong, L. Zhang, H. Farah, B. van Arem. All authors reviewed the results and approved the final version of the manuscript.

## Declaration of Conflicting Interests

The author(s) declared no potential conflicts of interest with respect to the research, authorship, and/or publication of this article.

## Funding

The author(s) disclosed receipt of the following financial support for the research, authorship, and/or publication of this article: This work was supported by Applied and Technical Sciences, a subdomain of the Dutch Institute for Scientific Research, through the project Safe and Efficient Operation of Automated and Human-Driven Vehicles in Mixed Traffic under Contract 17187. Additionally, partial support was provided by the Transport and Mobility Institute at Delft University of Technology.

## ORCID iDs

Yongqi Dong https://orcid.org/0000-0003-1159-9584
Haneen Farah https://orcid.org/0000-0002-2919-0253
Arkady Zgonnikov https://orcid.org/0000-0002-6593-6948
Bart van Arem https://orcid.org/0000-0001-8316-7794